\documentclass[11pt,a4paper]{article}
\usepackage[hyperref]{acl2020}
\usepackage{times}
\usepackage{latexsym}

\usepackage{microtype}

\usepackage{url}

\usepackage{mathrsfs}
\usepackage{graphicx}
\usepackage{courier}
\usepackage{amsmath}
\usepackage{tabularx}
\usepackage{float}
\usepackage{multicol}
\usepackage{microtype}

\aclfinalcopy 
\def\aclpaperid{218} 

\title{Parallel Corpus Filtering via Pre-trained Language Models}

\author{Boliang Zhang, Ajay Nagesh, and Kevin Knight \\
	    DiDi Labs \\
        {\tt \{boliangzhang, ajaynagesh, kevinknight\}@didiglobal.com}\\
        }
        
\date{}

\begin{document}
\maketitle
\begin{abstract}

Web-crawled data provides a good source of parallel corpora for training machine translation models. It is automatically obtained, but extremely noisy, and recent work shows that neural machine translation systems are more sensitive to noise than traditional statistical machine translation methods. In this paper, we propose a novel approach to filter out noisy sentence pairs from web-crawled corpora via pre-trained language models. We measure sentence parallelism by leveraging the multilingual capability of BERT and use the Generative Pre-training (GPT) language model as a domain filter to balance data domains.
We evaluate the proposed method on the WMT 2018 Parallel Corpus Filtering shared task, and on our own web-crawled Japanese-Chinese parallel corpus. Our method significantly outperforms baselines and achieves a new state-of-the-art. In an unsupervised setting, our method achieves comparable performance to the top-1 supervised method.
We also evaluate on a web-crawled Japanese-Chinese parallel corpus that we make publicly available.



\end{abstract}

\section{Introduction}

Training modern neural machine translation (NMT) systems requires large parallel-text resources. Publicly-available parallel corpora are mostly paired with English, such as German-English, French-English, Chinese-English, etc., and their domains are limited. 
For building machine translation systems between non-English language pairs, such as Chinese and Japanese, existing parallel corpora are insufficient and often low quality. To address this problem, system builders have trained NMT systems on web-crawled data and achieved promising results~\citep{xu2017zipporah,junczys2018dual,schwenk2018filtering,schwenk2019wikimatrix}. However, data automatically crawled from the web is extremely noisy. \citet{khayrallah2018impact} and \citet{belinkov2017synthetic} show that neural translation models are far more sensitive to noisy parallel training data than statistical machine translation. Data selection methods that can filter noisy parallel sentences from large-scale web crawled resources are in demand.


In this paper, we study the problem in a real-world scenario where we crawl a large Japanese-Chinese parallel corpus from various websites and build open-domain machine translation systems between Japanese and Chinese, by filtering the web crawled parallel corpus. In addition, a small amount of clean parallel data is available, in the software domain.  In order to confirm our results on a public data, we also apply our filter to the WMT 2018 German-English Parallel Corpus Filtering shared task.

Previous work on parallel corpus filtering performs poorly in our scenario as it either requires large clean parallel corpora or dictionaries~\cite{xu2017zipporah,artetxe2018margin,junczys2018dual,chaudhary2019low}, or relies on multilingual word embeddings and neglects context when measuring translation parallelism~\cite{hangya2018unsupervised}.


In this paper, we propose a simple but effective parallel corpus filtering method.
Multilingual BERT~\cite{devlin2018bert} projects multilingual sentences into a shared space and has shown a great potential for cross-lingual model transfer~\cite{pires2019multilingual}. We use pre-trained multilingual BERT as prior knowledge
and fine-tune it on a synthetic dataset. 
This multilingual BERT-based classifier forms an {\em acceptability filter} that determines whether or not a sentence pair consists of a bona-fide translation.

As the domain of training data largely affects machine translation model performance, we also introduce a domain filter. It uses the pre-trained Generative Pre-training (GPT) as in-domain language model and is an extension of the existing cross-entropy difference based domain filter~\cite{moore-lewis-2010-intelligent,junczys2018dual}.

We evaluate our proposed method on the WMT 2018 German-English Parallel Corpus Filtering shared task and achieve a new state-of-the-art. Our unsupervised method achieves comparable performance to the top system that is trained on millions of clean parallel sentence pairs. Our proposed methods also significantly outperform baselines in our own Japanese-Chinese parallel corpus filtering task. 

We make the following contributions:
\begin{itemize}
	\item We propose a novel approach to filter noisy parallel corpora by using pre-trained language models. Our approach outperforms strong baselines and achieves a new state-of-the-art.
	\item We devise an unsupervised filtering approach that does not require an identifiable clean subset of parallel segments.  Our unsupervised method matches the results of previous supervised methods.
	\item We release a large web-crawled Japanese-Chinese parallel corpus which can be a useful resource for machine translation research on non-English language pairs.\footnote{\url{http://iwslt.org/doku.php?id=open_domain_translation}}
\end{itemize}






\section{Related Work}
Several recent works address parallel corpus filtering.
\citet{denkowski2012cmu}, \citet{dyer2010cdec} and \citet{heafield2011kenlm} use language models and word alignments to determine how likely sentences are to be a good translation of another. \citet{xu2017zipporah} introduce a noise filtering tool, Zipporah, that discriminates parallel and non-parallel sentences based on word-frequency vectors and a dictionary. 
\citet{junczys2018dual} proposes a dual conditional cross-entropy filtering method, which achieved first place in the WMT 2018 German-English Parallel Corpus Filtering shared task. They train two translation models in inverse directions on millions of parallel sentences and score sentence pairs based on the word-normalized conditional cross-entropy from the translation models.
\citet{artetxe2018margin} and \citet{schwenk2018filtering} propose a margin-based scoring method that compares the similarity of the source and target sentence representations. The sentence representations are produced by a sentence encoder trained on clean parallel data via a neural encoder-decoder architecture. Other works based on sentence embeddings include \citet{hangya2018unsupervised} and \citet{littell2018measuring}, as well as \citet{schwenk2019wikimatrix}, which mines millions of parallel sentences in 1620 language pairs from Wikipedia. These encoder-decoder based methods require large amounts of clean parallel training data and are not applicable in our scenario where available data is noisy.
\citet{2020jazhsharedtask} organize an open domain translation challenge where participants are provided a large, noisy set of Japanese-Chinese segment pairs built from web data, and the task is to clean the noisy data and build an end-to-end machine translation system.



Work on data selection is also related. \citet{moore-lewis-2010-intelligent,junczys2018dual} select domain-related data by computing the cross-entropy difference between in-domain and out-domain language models. \citet{duh2013adaptation} use neural language models for data selection. 
\citet{axelrod2011domain} and \citet{axelrod2015class} expand cross-entropy difference filtering to both sides of the parallel corpus. Since we aim to build a general machine translation system, instead of selecting data that are relevant to a specific domain, we select data whose domains are as general as possible, by using Generative Pre-training (GPT) models trained on large and diverse corpora.

\section{Method}

In this section we introduce a language detection filter, a translation-acceptability filter, and a domain filter. Each filter produces a score 
for every candidate source/target sentence pair. The partial score produced by each filter ranges from 0 to 1. Values beyond this range are normalized by min-max normalization: $\hat{y}=(y-\textbf{min})/(\textbf{max}-\textbf{min})$.
The final score is the product of the partial scores.

\subsection{Language Detection Filter}


Targeting a web-crawler at a given language pair still results in many pages written in the wrong language.  For example, while a URL pair may clearly indicate translation (e.g., ``.jp'' and ``.zh''), it may happen that the text content is simply copied rather than translated.  We observe this in both our Japanese-Chinese data and the German-English Paracrawl data set. It is necessary to filter out sentence pairs with undesired languages.

We adopt the fastText~\cite{joulin2016bag,joulin2016fasttext} language identification toolkit in our language detection filter. For each sentence, the toolkit produces a list of language candidates and their corresponding confidence scores. We select the language that has the highest confidence score from fastText as the language of the sentence. Sentence pairs that have both of the elements detected as the desired language are assigned score \textbf{1} and otherwise \textbf{0}. By discarding sentence pairs with undesired language IDs, we filter out 27\% of our Chinese-Japanese parallel sentences and nearly 70\% of the German-English parallel sentences from Paracrawl data set.

\subsection{Acceptability Filter}
\label{section:acceptability_filter}

In this section, we introduce our translation acceptability filter, one of the main contributions in the paper.  It aims to measure the parallelism of sentence pairs and filter out sentence pairs that are not mutual translations. 

The pre-trained language model BERT~\cite{devlin2018bert} has been shown to be effective in many NLP tasks as it produces better and meaningful contextualized word representations.  Multilingual BERT, a transformer Masked Language Model pre-trained on Wikipedia dumps of 104 languages, shows remarkable multilingual capability, given that it is not exposed to any multilingual signals, such as parallel data or dictionaries. A thorough study by \citet{pires2019multilingual} shows the promising zero-shot cross-lingual model transfer ability of multilingual BERT on named entity recognition and part-of-speech tagging tasks. They hypothesize that having language-universal word pieces, such as numbers and URLs, mapped to a shared space forces the co-occurring pieces to also be mapped to a shared space, thus spreading the effect to other word pieces, until different languages are close in the shared space.

We use pre-trained multilingual BERT to encode a sentence pair $(s, t)$ and create the sentence embeddings $v_s$ and $v_t$ by using the representations of the [CLS] token of $s$ and $t$. We find that the cosine similarity between $v_s$ and $v_t$ does not necessarily reflect the parallelism of sentence $s$ and $t$. We suspect that the word representations from multilingual BERT are loosely aligned across languages as there is no parallel data or dictionary used during the pre-training. A similar observation was made in \citet{lample2017unsupervised}, where the cross-lingual word embeddings learned in an unsupervised manner are loosely aligned. However, after fine-tuning on a few anchor pairs (word translations), they become more aligned.

Similarly, we use an unsupervised synthetic training set as anchors to fine-tune multilingual BERT with a binary classification objective. \citet{xu2017zipporah} did similar work to train a filtering classifier on synthetic data, but via bag-of-words translation features.

\textbf{Synthetic Training Set.} In cases where a small number of clean parallel sentence pairs are available, we use them as positive training samples for our classifier. In Japanese-Chinese filtering, we use around 300k sentence pairs, mostly from open-source software documentation,\footnote{GNOME, Ubuntu, OpenOffice, and KDE data set, from \url{http://opus.nlpl.eu/}} as our positive samples.  In extreme cases where no identifiable, clean parallel data is available, we sub-select high quality parallel sentences, which are used as positive samples, from the noisy parallel corpus based on the Hunalign~\cite{varga2007parallel} sentence-alignment score.
We sample negative instances by simulating the noise produced by web crawling and alignment.
Given a positive pair $(s, t)$, we create a negative sample by randomly choosing one of the following options:
\begin{itemize}
	\item Randomly select a target sentence from its adjacent sentences within a window size of $k$ (where $k=2$ in our experiments).
	\item Randomly truncate 30\%-70\% of the source or target sentence.
	\item Swap the order of 30\%-70\% words of the source or target sentence.
\end{itemize}
To balance the training set, we create the same number of positive instances and sampled negative instances.

\textbf{Binary Classification Objective.} 
We feed the sentence pair $(s, t)$ into multilingual BERT, which accepts two-sentence input due to its next-sentence prediction objective~\cite{devlin2018bert}.  Instead of using the [CLS] token representation, we use a Convolutional Network (CNN) layer that takes the BERT output and generates the final representation of the pair. Our experiments show that using CNN layer pooling achieves marginal gains over [CLS] pooling. The final layer is a feed-forward network with a softmax activation function to produce label probabilities.  We use the softmax probability as the degree of parallelism.





\subsection{Domain Filter}

Web-crawled data contains noise of various types, due to the complicated structure of web pages. By inspecting the training data generated by the above methods, we notice much of the content is not well-formed, e.g., concatenated lists of months and dates, randomly mixed content from tables, series of emojis and punctuation marks, etc. These are certainly written in the desired language, thus not filtered out by language detection. The translation acceptability filter also accepts them.  However, such malformatted data is not helpful to machine translation models, and we prefer a training corpus to contain meaningful content.

For our domain filter, we adopt the cross-entropy difference scoring method proposed by~\citet{moore-lewis-2010-intelligent} and \citet{junczys2018dual}. More specifically, we treat a general domain monolingual corpus as our in-domain data set $\textbf{I}$, and the noisy parallel corpus without any filtering as our non-domain data set $\textbf{N}$. We train two language models $\textbf{L}_{I}$ and $\textbf{L}_{N}$ and measure how the target sentence $\textbf{t}$ is domain-related to $\textbf{I}$ and less domain-related to $\textbf{N}$ by a perplexity ratio, which is a transformation of cross-entropy difference:
\begin{displaymath}
\resizebox{.25 \textwidth}{!}{
$\hat{f}_{\texttt{dom}}(s, t)=\frac{\textbf{PPL}_{N}(t)}{\textbf{PPL}_{I}(t)}$
}
\label{eq:}
\end{displaymath}
where $\textbf{PPL}_{M}(x)$ is the word-normalized perplexity of the sentence $x$ defined by the language model $\textbf{L}_{M}$:
\begin{displaymath}
\resizebox{.45 \textwidth}{!}{
$\textbf{PPL}_{M}(x)=\exp(\frac{1}{|x|}\sum\limits_{i=1}^{|x|}\log P_{M}(x_i|x_{<i}))$
}
\end{displaymath}
The intuition is fairly straightforward: the higher the perplexity of the sentence to the non-domain corpus and the lower the perplexity of the sentence to the in-domain corpus, the more likely the sentence is meaningful.

Our contribution is to use GPT~\cite{radford2019language} as our in-domain language model, instead of news domain text~\cite{junczys2018dual}. This minor yet crucial change yields non-trivial performance gains in our experiments for German-English parallel corpus filtering. As GPT is trained on data from various sources, such as Wikipedia, Reddit, news websites, etc., it covers a wide range of domains, so our filtered data is more diverse and performs better on multi-domain test sets, as well as in the real world application.

For our in-domain language model, we use pre-trained Chinese GPT\footnote{https://github.com/dbiir/UER-py} for Japanese-Chinese and pre-trained GPT-2\footnote{https://github.com/huggingface/transformers} for German-English. We randomly sample 4 million sentences from the unfiltered noisy parallel corpus and use KenLM~\cite{heafield2011kenlm} to train the non-domain language model.  Perplexity scores from different language models are compatible.

Following~\citet{junczys2018dual}, we introduce two operations, clip and cutoff, to post-process the domain filter score $\hat{f}_{\texttt{dom}}(s, t)$. The clip operation clips the maximum value of the domain score to a threshold $\tau_{\texttt{clip}}$:

\begin{displaymath}
\resizebox{.36 \textwidth}{!}{
$f_{\texttt{clip}}(x, \tau_{\texttt{clip}})=\textbf{min}(x, \tau_{\texttt{clip}})$
}
\end{displaymath}
and the cutoff operation modifies scores below a threshold $\tau_{\texttt{cutoff}}$ and changes them to $0$:

\begin{displaymath}
\resizebox{.50 \textwidth}{!}{
$
f_{\texttt{cutoff}}(x, \tau_{\texttt{cutoff}})=
\begin{cases}
x, & \text{if } x > \tau_{\texttt{\texttt{cutoff}}}\\
0, & \text{otherwise}\\
\end{cases} 
$
}
\end{displaymath}
$\tau_{\texttt{clip}}$ prevents a high monolingual in-domain score from overwriting scores from other filters. $\tau_{\texttt{cutoff}}$ eliminates out-domain sentence pairs and ensures that highly parallel sentence pairs are at least somewhat in-domain. We tune $\tau_{\texttt{clip}}$ and $\tau_{\texttt{cutoff}}$ on the development set.

The scoring method of our final domain filter becomes:
\begin{displaymath}
\resizebox{.5 \textwidth}{!}{
$f_{\texttt{dom}}(s, t)=f_{\texttt{clip}}(f_{\texttt{cutoff}}(\hat{f}_{\texttt{dom}}(s, t), \tau_{\texttt{cutoff}}), \tau_{\texttt{clip}})$
}
\end{displaymath}



\section{Experiments and Results}

\subsection{WMT 2018 Parallel Corpus Filtering}

We use the WMT 2018 Parallel Corpus Filtering shared task~\cite{koehn2018findings} as a benchmark to evaluate our methods. Participants in the shared task are provided a very noisy 1~billion word (English token count) German-English corpus crawled from the web by the Paracrawl project.\footnote{\url{https://paracrawl.eu}} The task is to sub-select clean sentence pairs amounting to (a) 10 million words, and (b) 100 million words, counted on the English side. The quality of the resulting subsets is determined by training a neural machine translation system (Marian)\footnote{\url{https://github.com/marian-nmt/marian} (We do not evaluate our method using Moses, the statistical machine translation system provided by WMT, as neural machine translation better fits our real world scenario.)}~\cite{mariannmt} on this data. The quality of the machine translation system is measured by BLEU score on six test sets from various domains. As the task is to address the challenge of the data quality and not domain-relatedness of the data for a particular use, sub-sampling the corpus for relevance to the news domain is not encouraged by the shared task organizers.
All parameters used for training Marian machine translation models are the same as described in~\citet{koehn2018findings}.
We use $\texttt{CLIP}=5$ and $\texttt{CUTOFF}=1.5$ in the experiments. We use 4 GPUs for training.

\subsection{Web-Crawled Japanese-Chinese Parallel Corpus Filtering}

Due to the lack of publicly available Japanese-Chinese parallel corpus, we build a data harvesting pipeline to fetch Japanese-Chinese parallel text from the Internet. The crawled bi-text are extremely noisy, but we rely on the proposed parallel corpus filtering method to clean up the data and eventually train a satisfactory machine translation system. In this paper, we use these crawled data as another test bed to evaluate our proposed method.

A single run of the 
of the data harvesting pipeline is the following. We first identify Japanese-Chinese parallel webpages by programmatically analyzing the URL structure of the 5 billion URLs from CommonCrawl,\footnote{\url{https://commoncrawl.org/}} for example, \url{https://www.gotokyo.org/jp/} and \url{https://www.gotokyo.org/cn/} only differ by \url{jp} and \url{cn}. Then we download the webpages and conduct a series of cascaded data cleaning methods, including removing HTML markups, sentence segmentation, etc. Finally we perform segment alignment and filtering. 
Our workflow consists of several runs of the data harvesting pipeline with entry points at different modules (for instance, a more targeted crawling of higher quality material from a previous run). 




We also integrate existing Japanese-Chinese parallel datasets from other publicly available sources for a final parallel data size of 527m characters in 20.9M parallel segments. 

We include all details of our data harvesting pipeline, as well as the statistics of the obtained dataset, in Appendix~\ref{sec:appendix}.


\textbf{Test and Development Dataset.} 
We curate two  parallel test sets by manually processing web data involving daily expressions (337 parallel segments) and news (437 parallel segments). For our development set, we use 5304 Japanese-Chinese basic expressions.

\subsection{Results and Analysis}

\begin{table*}  
    \centering
    \begin{tabular}{lccrr}
        \textbf{Method} & \textbf{Supervised} & \textbf{Unsupervised} & \textbf{10M} & \textbf{100M} \\ \hline
        \citet{junczys2018dual} {\small \textit{top-1}} & x & & 28.62 & 32.05 \\
        \citet{lu2018alibaba} {\small \textit{top-2}} & x & & 27.60 & 31.93 \\
        \citet{lo2018accurate} {\small \textit{top-3}} & x & & 27.41 & 31.88 \\
        \citet{hangya2018unsupervised} &  & x & 22.96 & 30.54 \\ 
        \citet{chaudhary2019low} & x & & 26.98 & 30.77 \\ \hline
        \textbf{adequacy} (our replication of J-D 2018)& x &   & 27.12 & 31.20 \\
        \hspace{2mm} + domain-news (our replication of J-D 2018) & x &   & 28.66 & 32.01 \\
        \hspace{2mm} + domain-GPT & x &   & $^{\dagger}$\textbf{29.09} & $^{\dagger}$\textbf{32.11} \\ \hline
        \textbf{supervised acceptability} & x &   & 27.09 & 31.56 \\
        \hspace{2mm} + domain-GPT & x &   & 28.94 & 32.03 \\ \hline
        \textbf{unsupervised acceptability} &  & x & 27.03 & 30.65 \\
        \hspace{2mm} + domain-GPT &  & x  & $^{\ddagger}$\underline{28.68} &  $^{\ddagger}$\underline{32.02} \\ \hline
         \multicolumn{5}{p{0.9\textwidth}}{\small - all methods above apply language detection filter beforehand.}\\
        \multicolumn{5}{p{0.9\textwidth}}{\small $\dagger$ our new state-of-the-art combines \textit{adequacy}~\cite{junczys2018dual} + our proposed \textit{domain-GPT}.}\\
        \multicolumn{5}{p{0.9\textwidth}}{\small $\ddagger$ our \textit{unsupervised acceptability} + \textit{domain-GPT} is comparable to top supervised method.}\\
    \end{tabular}
    \caption{BLEU scores of German-English neural MT systems trained on 10 million and 100 million word training data selected by different methods. The scores are averaged BLEU scores across the six test sets from WMT 2018 parallel corpus filtering task. {\em domain-news} trains an in-domain language model on news corpus, while {\em domain-GPT} uses the pre-trained GPT language model.}
    \label{tab:bleu_scores}
\end{table*}

\begin{table*}[h]
    \centering
    \begin{tabular}{lrrrr}
        \textbf{Methods} & \textbf{JA-ZH} &  $\textbf{\%}^*$ & \textbf{ZH-JA} & $\textbf{\%}^*$  \\
        \hline
        unfiltered & 22.92 & 100 & 22.27 & 100 \\
        \citet{chaudhary2019low} & 23.46 & 75 & 26.22  & 70 \\
        \hline 
        \textbf{adequacy} (our replication of J-D 2018) & 23.91 & 90 & 24.51 & 90 \\
        \hspace{2mm} + domain-GPT & 24.00 & 65 & - & - \\
        \hline
        \textbf{acceptability} & \textbf{25.53} & 75 &  \textbf{28.54} & 50 \\
        \hspace{2mm} + domain-GPT  & 25.49 & 50 & - & - \\
        \hline
        \multicolumn{4}{p{0.7\textwidth}}{\small - all methods above apply language detection filter beforehand.}\\
        \multicolumn{4}{p{0.7\textwidth}}{\small * percentage of raw parallel sentences used for MT training.}\\
    \end{tabular}
    \caption{BLEU scores of Japanese-Chinese and Chinese-Japanese MT systems trained on data sets generated by various filtering methods. We rank sentence pairs by filtering scores and train an MT system on $N$ percent of the top ranked data. $N$ is selected based on the development set and we report the best BLEU score. {\em domain-GPT} is the domain filter whose in-domain language model is the pre-trained GPT language model; note that for ZH-JA, we do not have access to pre-trained Japanese GPT.}
    \label{tab:ja-zh}
\end{table*}

\textbf{WMT 2018 Parallel Corpus Filtering.} 
Table~\ref{tab:bleu_scores} presents the BLEU scores of neural machine translation systems trained on 10 million and 100 million words of training data, selected by different filtering methods.
In the table, we list the top three performers from the shared task, as well as another two work that are similar to ours.
\citet{junczys2018dual} has a dual conditional cross-entropy adequacy filter and a domain filter trained on news corpora.
\citet{hangya2018unsupervised} generate sentence embeddings by using unsupervised word embedding alignment and measure parallelism via multilingual sentence embedding similarity.
\citet{chaudhary2019low} leverage massive publicly available English-German parallel corpora to train multilingual sentence embeddings via bidirectional Long Short Term Memory (LSTM) encoder-decoder network.


We replicate the adequacy and domain-news filters from \citet{junczys2018dual} and obtain similar results. By replacing the domain-news filter with our domain-GPT filter, we achieve new state-of-the-art scores on 10M and 100M word data sets (bold scores in the table). Given the very compact score range in the shared task~\cite{koehn2018findings}, we consider this gain very successful.
It is stated in the shared task that the test sets are from multiple domains. Domain-news filter in \citet{junczys2018dual} tends to select sentence pairs from news domain as the filter is trained on news domain data, and this leads to a biased parallel corpus for training machine translation system. Our proposed domain-GPT filter is trained from various sources and thus covers a wide range of domains, so our filtered data is more diverse and performs better on multi-domain test sets.

For our supervised acceptability filter, we train a mulitlingual BERT classifier on clean parallel sentences as positive examples and randomly sampling negative instances, using the method described in Section~\ref{section:acceptability_filter}.
For our unsupervised acceptability filter, we rank noisy parallel sentences by (a) the alignment score from Hunalign, and (b) the GPT domain filter score.  We then select the top 10M words (counted on English side) worth of sentence pairs as positive examples. This makes the method completely unsupervised, not requiring any identifiable clean parallel data. With finetuning multilingual BERT on sentences pairs aligned by Hunalign, the unsupervised acceptability already achieves comparable performance to \citet{chaudhary2019low} which use massive public parallel data. After applying the unsupervised domain-GPT filter, we achieve a surprisingly good result (underlined scores in the table), comparable to the best supervised method.

\textbf{Japanese-Chinese Parallel Corpus Filtering.} In Table~\ref{tab:ja-zh}, we evaluate machine translation systems trained on data generated by different filtering methods.  {\em Unfiltered} refers to data generated by Hunalign without any filtering. \citet{chaudhary2019low} refer to LASER, the top performing filtering system in WMT 2019 Parallel Corpus Filtering shared task. We use the pre-trained 93-language LASER model to generate sentence pair scores. The model is trained on a large parallel corpus that contains 3.2M English-Japanese and 8.2M English-Chinese sentence pairs (English is used as pivot to connect Japanese and Chinese during their training). {\em Adequacy} refers to the dual conditional cross-entropy filtering method that we replicate from~\citet{junczys2018dual}. It is trained on around 300k high quality software-domain parallel sentences from Microsoft Developer Network (MSDN) and Ubuntu. The GPT domain filter uses a pre-trained Chinese GPT\footnote{pre-trained Mixedlarge corpus + GptEncoder + LmTarget Model in \url{https://github.com/dbiir/UER-py}} as the in-domain language model and trains a four-gram KenLM~\cite{heafield2011kenlm} language model on the Chinese side of our 4~million unfiltered noisy parallel sentences as a non-domain language model. {\em Acceptability} is our proposed multilingual BERT based filtering method, which is trained on a synthetic dataset, where we use 300k high-quality software domain parallel sentences as positive examples and sample equal-sized negative sentence pairs, using the sampling methods described in Section~\ref{section:acceptability_filter}. 

\citet{chaudhary2019low} train a multilingual sentence encoder on various English-Foreign\_Language parallel corpus and prove the zero-shot cross-lingual transfer capability between non-English pairs, such as Japanese and Chinese. 
However, when English is used as the pivot, the distance between Japanese and Chinese become larger, resulting in not effectively capturing the correlation between them.
The conditional cross-entropy metric in {\em adequacy} relies on the quality of machine translation system. Due to the difficulty of training high-quality machine translation systems on 300k sentence pairs, the adequacy filter cannot produce accurate conditional cross-entropy. The GPT domain filter assigns higher score to sentences that are more like human natural language and downgrades malformatted sentence pairs. It is effective in the German-English filtering task, where a fixed-size subset is selected and we want to fill the subset with as much domain relevant data as possible. However, to best fit the real world scenario where the goal is to have the best machine translation system, we do not limit the amount of data to select for training machine translation system and let the system decide the amount of the data to select, according to each filtering method. 
We rank sentence pairs by their filtering scores and train a MT system on $N$ percentage of the top ranked data. $N$ is selected based on the development set and we report the best BLEU score.
Under this setting, adding a domain filter makes the model use less data ($N=50\%$ vs $N=75\%$), but we do not observe any performance gain, as we suspect that the malformatted but parallel sentence pairs are neither harmful or helpful to the model, and filtering them out makes no difference in performance of the model.


\textbf{High Precision Parallel Corpus Filtering.} 
\begin{figure}
    \centering
    \includegraphics[width=7.5cm]{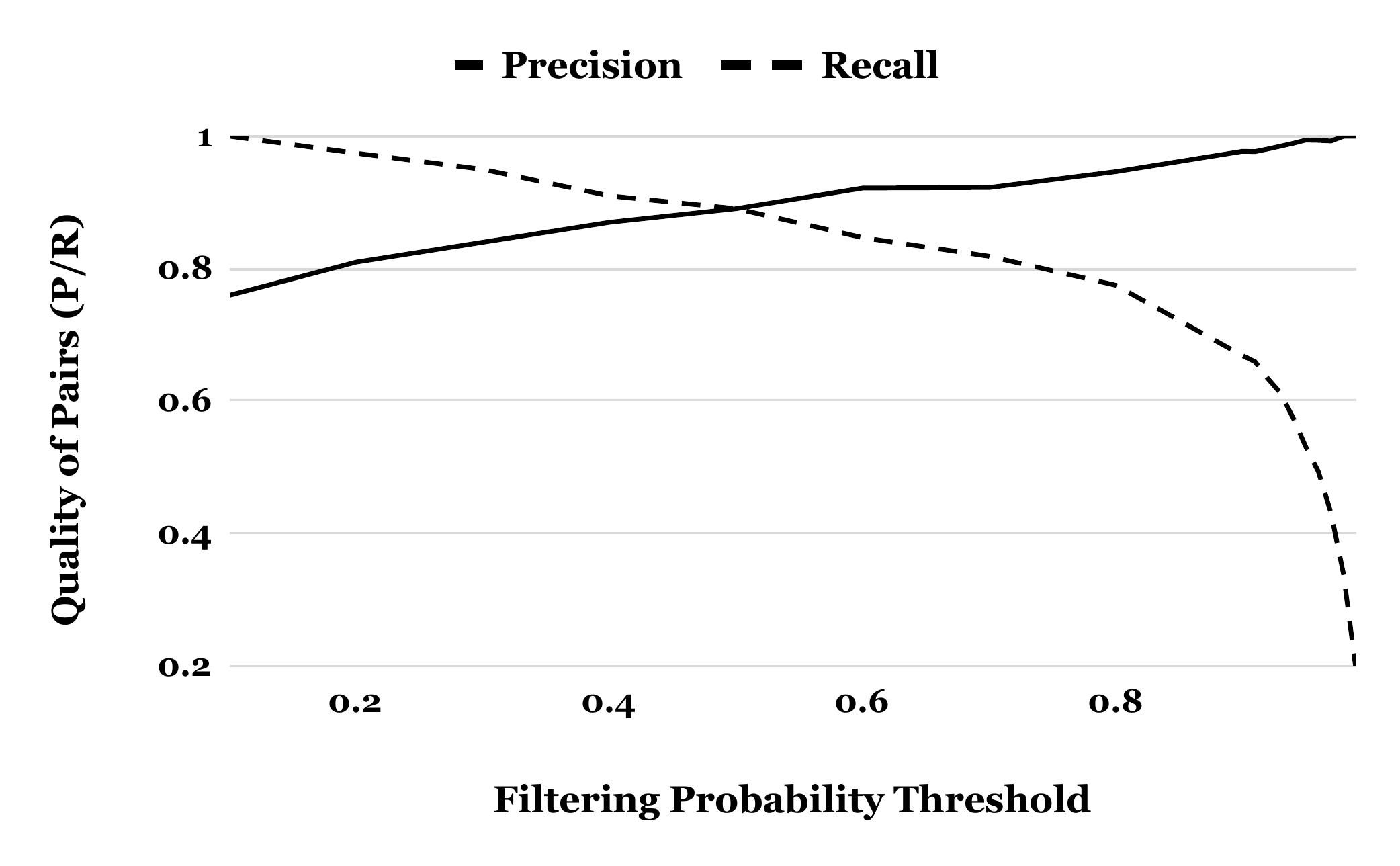}
    \caption{Precision and recall curves of the acceptability filter on our internal JA-ZH filtering test set. The threshold is based on the classifier probability produced by the softmax layer. When threshold set to 0.9, we obtain 97.7\% precision parallel sentence pairs at 66.9\% recall.}
    \label{fig:filtering_acc_pr_curve}
\end{figure}
For analysis purposes, we manually annotate a small set of 320 sentence pairs randomly selected from our original web crawled Japanese-Chinese data set. 
24\% of the sentence pairs are labeled ``not mutual translations.''  
As stated in \citet{khayrallah2018impact}, neural machine translation models are more sensitive to noise than statistical machine translation models, so having high precision filtering results as training data is necessary.
In Figure~\ref{fig:filtering_acc_pr_curve}, we show precision and recall curves for our proposed filtering method on this labeled test set, under different threshold settings. The threshold is selected based on the filtering classifier probability produced by the softmax layer. 
By setting the threshold to 0.9, we are able to obtain 97.7\% precision high-quality parallel sentences, while still having 66.9\% recall.

\section{Conclusions}

In this paper, we address the parallel corpus filtering problem in machine translation. We propose a novel filtering method using pre-trained language models. Our method outperforms strong baselines and achieves a new state-of-the-art. We release a large Japanese-Chinese web crawled parallel corpus for the research purposes. Because it is artificial to use synthetic data for training a filter classifier, future work can focus on a better objective that models parallelism more smoothly. Future work also includes extending the method to low-resource languages not covered by multilingual BERT.



\section*{Acknowledgments}
We would like to thank the anonymous reviewers for their constructive feedback. 

\bibliography{acl2020}

\begin{thebibliography}{33}
\expandafter\ifx\csname natexlab\endcsname\relax\def\natexlab#1{#1}\fi

\bibitem[{Artetxe and Schwenk(2019)}]{artetxe2018margin}
Mikel Artetxe and Holger Schwenk. 2019.
\newblock Margin-based parallel corpus mining with multilingual sentence
  embeddings.
\newblock In \emph{Proceedings of the Annual Meeting of the Association for
  Computational Linguistics}.

\bibitem[{Axelrod et~al.(2011)Axelrod, He, and Gao}]{axelrod2011domain}
Amittai Axelrod, Xiaodong He, and Jianfeng Gao. 2011.
\newblock Domain adaptation via pseudo in-domain data selection.
\newblock In \emph{Proceedings of the Conference on Empirical Methods in
  Natural Language Processing (EMNLP)}.

\bibitem[{Axelrod et~al.(2015)Axelrod, Vyas, Martindale, and
  Carpuat}]{axelrod2015class}
Amittai Axelrod, Yogarshi Vyas, Marianna Martindale, and Marine Carpuat. 2015.
\newblock Class-based n-gram language difference models for data selection.
\newblock In \emph{Proceedings of the International Workshop on Spoken Language
  Translation (IWSLT)}.

\bibitem[{Belinkov and Bisk(2018)}]{belinkov2017synthetic}
Yonatan Belinkov and Yonatan Bisk. 2018.
\newblock Synthetic and natural noise both break neural machine translation.
\newblock In \emph{Proceedings of the Sixth International Conference on
  Learning Representations (ICLR)}.

\bibitem[{Chaudhary et~al.(2019)Chaudhary, Tang, Guzm{\'a}n, Schwenk, and
  Koehn}]{chaudhary2019low}
Vishrav Chaudhary, Yuqing Tang, Francisco Guzm{\'a}n, Holger Schwenk, and
  Philipp Koehn. 2019.
\newblock Low-resource corpus filtering using multilingual sentence embeddings.
\newblock In \emph{Proceedings of the Fourth Conference on Machine
  Translation}.

\bibitem[{Chu et~al.(2015)Chu, Nakazawa, and Kurohashi}]{wikipedia_corpus}
Chenhui Chu, Toshiaki Nakazawa, and Sadao Kurohashi. 2015.
\newblock Integrated parallel sentence and fragment extraction from comparable
  corpora: A case study on {Chinese--Japanese Wikipedia}.
\newblock \emph{ACM Transactions on Asian and Low-Resource Language Information
  Processing}.

\bibitem[{Dabre and Kurohashi(2017)}]{ted_corpus}
Raj Dabre and Sadao Kurohashi. 2017.
\newblock \href {http://arxiv.org/abs/1710.01025} {{MMCR4NLP:} multilingual
  multiway corpora repository for natural language processing}.
\newblock \emph{CoRR}, abs/1710.01025.

\bibitem[{Denkowski et~al.(2012)Denkowski, Hanneman, and
  Lavie}]{denkowski2012cmu}
Michael Denkowski, Greg Hanneman, and Alon Lavie. 2012.
\newblock The {CMU-Avenue French-English} translation system.
\newblock In \emph{Proceedings of the Seventh Workshop on Statistical Machine
  Translation}.

\bibitem[{Devlin et~al.(2019)Devlin, Chang, Lee, and
  Toutanova}]{devlin2018bert}
Jacob Devlin, Ming-Wei Chang, Kenton Lee, and Kristina Toutanova. 2019.
\newblock {BERT}: Pre-training of deep bidirectional transformers for language
  understanding.
\newblock In \emph{Proceedings of the Conference of the North {A}merican
  Chapter of the Association for Computational Linguistics: Human Language
  Technologies (NAACL-HLT)}.

\bibitem[{Duh et~al.(2013)Duh, Neubig, Sudoh, and Tsukada}]{duh2013adaptation}
Kevin Duh, Graham Neubig, Katsuhito Sudoh, and Hajime Tsukada. 2013.
\newblock Adaptation data selection using neural language models: Experiments
  in machine translation.
\newblock In \emph{Proceedings of the Annual Meeting of the Association for
  Computational Linguistics (ACL)}.

\bibitem[{Dyer et~al.(2010)Dyer, Weese, Setiawan, Lopez, Ture, Eidelman,
  Ganitkevitch, Blunsom, and Resnik}]{dyer2010cdec}
Chris Dyer, Jonathan Weese, Hendra Setiawan, Adam Lopez, Ferhan Ture, Vladimir
  Eidelman, Juri Ganitkevitch, Phil Blunsom, and Philip Resnik. 2010.
\newblock cdec: A decoder, alignment, and learning framework for finite-state
  and context-free translation models.
\newblock In \emph{Proceedings of the Annual Meeting of the Association for
  Computational Linguistics (ACL), System Demonstrations}.

\bibitem[{Hangya and Fraser(2018)}]{hangya2018unsupervised}
Viktor Hangya and Alexander Fraser. 2018.
\newblock An unsupervised system for parallel corpus filtering.
\newblock In \emph{Proceedings of the Third Conference on Machine Translation:
  Shared Task Papers}.

\bibitem[{Heafield(2011)}]{heafield2011kenlm}
Kenneth Heafield. 2011.
\newblock {KenLM}: Faster and smaller language model queries.
\newblock In \emph{Proceedings of the Sixth Workshop on Statistical Machine
  Translation}.

\bibitem[{Joulin et~al.(2016)Joulin, Grave, Bojanowski, Douze, J{\'e}gou, and
  Mikolov}]{joulin2016fasttext}
Armand Joulin, Edouard Grave, Piotr Bojanowski, Matthijs Douze, H{\'e}rve
  J{\'e}gou, and Tomas Mikolov. 2016.
\newblock {FastText.zip}: Compressing text classification models.
\newblock \emph{arXiv preprint arXiv:1612.03651}.

\bibitem[{Joulin et~al.(2017)Joulin, Grave, Bojanowski, and
  Mikolov}]{joulin2016bag}
Armand Joulin, Edouard Grave, Piotr Bojanowski, and Tomas Mikolov. 2017.
\newblock Bag of tricks for efficient text classification.
\newblock In \emph{Proceedings of the Conference of the {E}uropean Chapter of
  the Association for Computational Linguistics (EACL)}.

\bibitem[{Junczys-Dowmunt(2018)}]{junczys2018dual}
Marcin Junczys-Dowmunt. 2018.
\newblock Dual conditional cross-entropy filtering of noisy parallel corpora.
\newblock In \emph{Proceedings of the Third Conference on Machine Translation:
  Shared Task Papers}.

\bibitem[{Junczys-Dowmunt et~al.(2018)Junczys-Dowmunt, Grundkiewicz, Dwojak,
  Hoang, Heafield, Neckermann, Seide, Germann, Fikri~Aji, Bogoychev, Martins,
  and Birch}]{mariannmt}
Marcin Junczys-Dowmunt, Roman Grundkiewicz, Tomasz Dwojak, Hieu Hoang, Kenneth
  Heafield, Tom Neckermann, Frank Seide, Ulrich Germann, Alham Fikri~Aji,
  Nikolay Bogoychev, Andr\'{e} F.~T. Martins, and Alexandra Birch. 2018.
\newblock Marian: Fast neural machine translation in {C++}.
\newblock In \emph{Proceedings of the Annual Meeting of the Association for
  Computational Linguistics (ACL), System Demonstrations}.

\bibitem[{Khayrallah and Koehn(2018)}]{khayrallah2018impact}
Huda Khayrallah and Philipp Koehn. 2018.
\newblock On the impact of various types of noise on neural machine
  translation.
\newblock In \emph{Proceedings of the Workshop on Neural Machine Translation
  and Generation}.

\bibitem[{Koehn et~al.(2018)Koehn, Khayrallah, Heafield, and
  Forcada}]{koehn2018findings}
Philipp Koehn, Huda Khayrallah, Kenneth Heafield, and Mikel~L Forcada. 2018.
\newblock Findings of the {WMT} 2018 shared task on parallel corpus filtering.
\newblock In \emph{Proceedings of the Third Conference on Machine Translation:
  Shared Task Papers}.

\bibitem[{Lample et~al.(2018)Lample, Conneau, Denoyer, and
  Ranzato}]{lample2017unsupervised}
Guillaume Lample, Alexis Conneau, Ludovic Denoyer, and Marc'Aurelio Ranzato.
  2018.
\newblock Unsupervised machine translation using monolingual corpora only.
\newblock In \emph{Proceedings of the 6th International Conference on Learning
  Representations (ICLR)}.

\bibitem[{Lison and Tiedemann(2016)}]{opus_2016}
Pierre Lison and J{\"{o}}rg Tiedemann. 2016.
\newblock {OpenSubtitles2016}: Extracting large parallel corpora from movie and
  {TV} subtitles.
\newblock In \emph{Proceedings of the Tenth International Conference on
  Language Resources and Evaluation (LREC)}.

\bibitem[{Littell et~al.(2018)Littell, Larkin, Stewart, Simard, Goutte, and
  Lo}]{littell2018measuring}
Patrick Littell, Samuel Larkin, Darlene Stewart, Michel Simard, Cyril Goutte,
  and Chi-kiu Lo. 2018.
\newblock Measuring sentence parallelism using {Mahalanobis} distances: The
  {NRC} unsupervised submissions to the {WMT18} parallel corpus filtering
  shared task.
\newblock In \emph{Proceedings of the Third Conference on Machine Translation:
  Shared Task Papers}.

\bibitem[{Lo et~al.(2018)Lo, Simard, Stewart, Larkin, Goutte, and
  Littell}]{lo2018accurate}
Chi-kiu Lo, Michel Simard, Darlene Stewart, Samuel Larkin, Cyril Goutte, and
  Patrick Littell. 2018.
\newblock Accurate semantic textual similarity for cleaning noisy parallel
  corpora using semantic machine translation evaluation metric: The {NRC}
  supervised submissions to the parallel corpus filtering task.
\newblock In \emph{Proceedings of the Third Conference on Machine Translation:
  Shared Task Papers}.

\bibitem[{Lu et~al.(2018)Lu, Lv, Shi, and Chen}]{lu2018alibaba}
Jun Lu, Xiaoyu Lv, Yangbin Shi, and Boxing Chen. 2018.
\newblock Alibaba submission to the {WMT18} parallel corpus filtering task.
\newblock In \emph{Proceedings of the Third Conference on Machine Translation:
  Shared Task Papers}.

\bibitem[{Moore and Lewis(2010)}]{moore-lewis-2010-intelligent}
Robert~C. Moore and William Lewis. 2010.
\newblock Intelligent selection of language model training data.
\newblock In \emph{Proceedings of the Annual Meeting of the Association for
  Computational Linguistics (ACL)}.

\bibitem[{Ondrej~Bojar(2020)}]{2020jazhsharedtask}
Christian Federmann Jiatao Gu Fei Huang Ajay Nagesh Jan Niehues Elizabeth
  Salesky Sebastian St ̈uker Marco~Turchi Ondrej~Bojar, Marcello~Federico.
  2020.
\newblock Findings of the iwslt 2020 evaluation campaign.
\newblock In \emph{Proceedings of the 2020 International Conference on Spoken
  Language Translation (IWSLT)}.

\bibitem[{Pires et~al.(2019)Pires, Schlinger, and
  Garrette}]{pires2019multilingual}
Telmo Pires, Eva Schlinger, and Dan Garrette. 2019.
\newblock How multilingual is multilingual {BERT}?
\newblock In \emph{Proceedings of the Annual Meeting of the Association for
  Computational Linguistics (ACL)}.

\bibitem[{Radford et~al.(2019)Radford, Wu, Child, Luan, Amodei, and
  Sutskever}]{radford2019language}
Alec Radford, Jeff Wu, Rewon Child, David Luan, Dario Amodei, and Ilya
  Sutskever. 2019.
\newblock Language models are unsupervised multitask learners.

\bibitem[{Schwenk(2018)}]{schwenk2018filtering}
Holger Schwenk. 2018.
\newblock Filtering and mining parallel data in a joint multilingual space.
\newblock In \emph{Proceedings of the Annual Meeting of the Association for
  Computational Linguistics (ACL)}.

\bibitem[{Schwenk et~al.(2019)Schwenk, Chaudhary, Sun, Gong, and
  Guzm{\'a}n}]{schwenk2019wikimatrix}
Holger Schwenk, Vishrav Chaudhary, Shuo Sun, Hongyu Gong, and Francisco
  Guzm{\'a}n. 2019.
\newblock Wikimatrix: Mining 135m parallel sentences in 1620 language pairs
  from {Wikipedia}.
\newblock \emph{arXiv preprint arXiv:1907.05791}.

\bibitem[{Tiedemann(2012)}]{opus_2012}
J\"{o}rg Tiedemann. 2012.
\newblock Parallel data, tools and interfaces in {OPUS}.
\newblock In \emph{Proceedings of the Eight International Conference on
  Language Resources and Evaluation (LREC)}.

\bibitem[{Varga et~al.(2007)Varga, Hal{\'a}csy, Kornai, Nagy, N{\'e}meth, and
  Tr{\'o}n}]{varga2007parallel}
D{\'a}niel Varga, P{\'e}ter Hal{\'a}csy, Andr{\'a}s Kornai, Viktor Nagy,
  L{\'a}szl{\'o} N{\'e}meth, and Viktor Tr{\'o}n. 2007.
\newblock Parallel corpora for medium density languages.
\newblock \emph{Amsterdam Studies in the Theory and History of Linguistic
  Science}.

\bibitem[{Xu and Koehn(2017)}]{xu2017zipporah}
Hainan Xu and Philipp Koehn. 2017.
\newblock Zipporah: a fast and scalable data cleaning system for noisy
  web-crawled parallel corpora.
\newblock In \emph{Proceedings of the Conference on Empirical Methods in
  Natural Language Processing (EMNLP)}.

\end{thebibliography}
\bibliographystyle{acl_natbib}

\clearpage
\newpage
\appendix
\onecolumn
\section{Web-Crawled Parallel Data for Japanese-Chinese}
\label{sec:appendix}

\begin{figure*}[h]
    \centering
    \includegraphics[width=15cm, height=5cm]{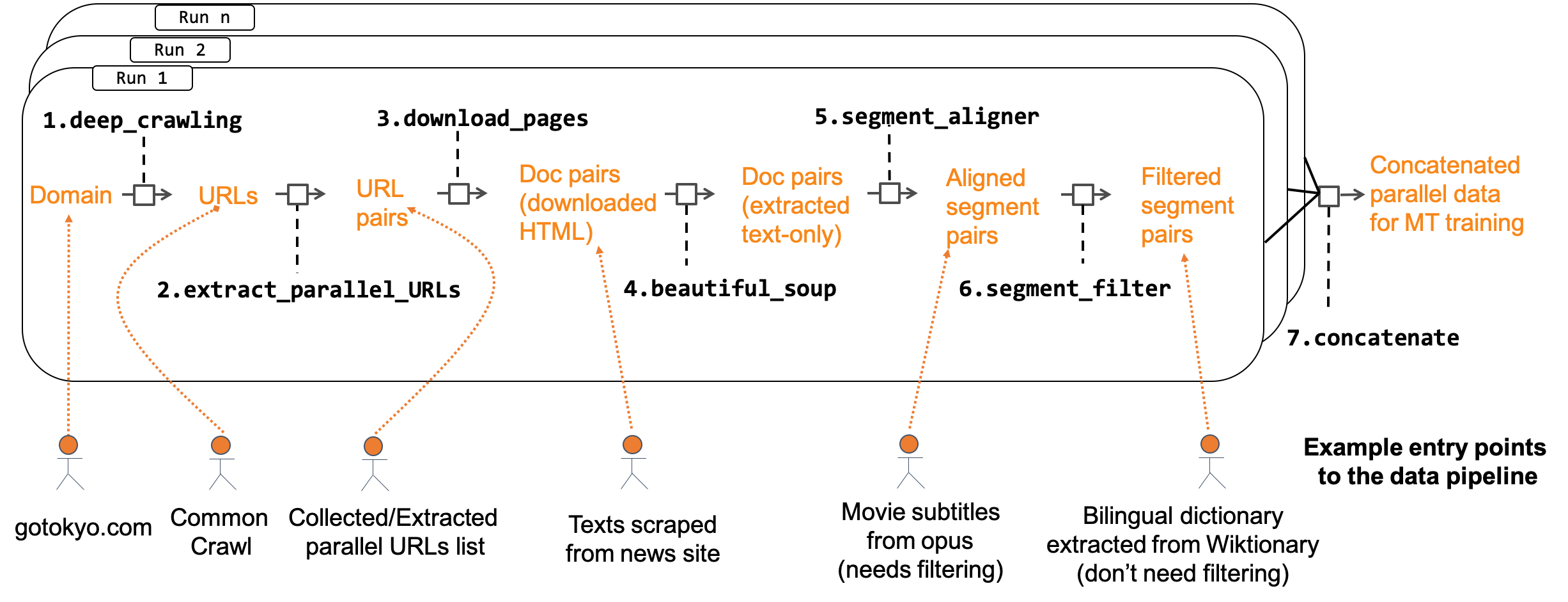}
    \caption{Our Japanese-Chinese parallel data harvesting pipeline. It consists of several modules, each of them numbered. The inputs to and outputs from each module are depicted in orange. The example entry points to the data pipeline are shown at the bottom of the diagram.}
    \label{fig:data_pipeline}
\end{figure*}

\begin{table*}[h]
    \centering
    \begin{tabular}{lrrl}
    \textbf{Source}  & \textbf{\# Segment-pairs} & \textbf{\# Characters (zh side)} & \textbf{Reference}\\
    \hline
    Web-crawled (pipeline) & 18,966,595 & 493,902,539 & - \\
    \hline 
    Linux documentation      & 92,250 & 1,549,964 & \citet{opus_2012}\\
    Open Subtitiles & 914,355 & 10,932,722 & \citet{opus_2016}\\
    TED & 376,441 & 5,345,867 & \citet{ted_corpus}\\
    Global Voices & 16,848 & 337,194 & \citet{opus_2012}\\
    Wikipedia & 228,565 & 5,067,489 & \citet{wikipedia_corpus}\\
    Wiktionary & 62,557 & 222,562 & wiktionary.org \\
    News Commentary & 570 & 65,038 & \citet{opus_2012}\\
    Tatoeba & 4,243 & 50,846  & tatoeba.org \\ 
    Facebook & 267,409 & 9,950,657 & \citet{schwenk2019wikimatrix}\\
    \hline

    \textbf{Total} & \textbf{20,929,833} & \textbf{527,424,878} & -
    \end{tabular}
    \caption{Japanese-Chinese parallel data assembled for our experiments.}
    \label{tab:data_statistics}
\end{table*}

This appendix describes our pipeline to extract parallel Japanese-Chinese parallel sentence fragments from the Internet (Figure~\ref{fig:data_pipeline}).  We start with  5~billion URLs from CommonCrawl.\footnote{\url{https://commoncrawl.org/}} We identify Japanese-Chinese parallel webpages by looking at URL structure (step~2). For example, \url{https://www.gotokyo.org/jp/} and \url{https://www.gotokyo.org/cn/} only differ by \url{jp} and \url{cn}. We download these potentially parallel page pairs (step~3), remove HTML and other markup metadata (step~4),\footnote{Using Python module \texttt{BeautifulSoup}} and split into sentence segments. We use off-the-shelf  Hunalign\footnote{http://mokk.bme.hu/en/resources/hunalign/} for segment alignment (step~5). We filter segment pairs by rough language ID and length ratio (step~6). We obtain 227k~URL pairs, 1.4m~segment pairs, and 28.7m characters of parallel data (measured on the Chinese side).

From the 227k~URL pairs above, we trace which site pairs yielded the most parallel data.  We then run a deep-crawling module on each of the 6000 most-promising sites,\footnote{Using the Python-based \texttt{scrapy} tool} and we process the resulting URLs using the rest of the pipeline.  Concatenating parallel data from all runs (step~7) and running a simple post-processing filter to remove objectionable content in the text gathered, we obtain around 494m characters of parallel data (measured on the Chinese side). 

We also integrate existing Japanese-Chinese parallel datasets from other publicly available sources for a final parallel data size 527m characters in 20.9m parallel segments. Table~\ref{tab:data_statistics} describes the various components of this dataset.

\end{document}